\documentclass[sigconf]{acmart}

\usepackage{enumitem}
\usepackage{amsmath}
\usepackage{graphicx}
\usepackage{enumitem}
\usepackage{multirow} 
\usepackage{fancyhdr}
\usepackage{caption}
\usepackage{array}
\usepackage{xcolor}
\newcolumntype{B}{>{\columncolor{blue!10}\bfseries\color{blue}}c}
\usepackage{bm}
\usepackage{colortbl}
\usepackage{threeparttable}
\usepackage{afterpage}
\usepackage{tablefootnote}
\usepackage{amsfonts}
\usepackage[normalem]{ulem}
\useunder{\uline}{\ul}{}

\usepackage{tikz}
\usepackage{pgfplots}

\definecolor{color1}{RGB}{247,216,213}  
\definecolor{color2}{RGB}{222,235,242}  
\definecolor{color3}{RGB}{250,230,203}  
\definecolor{color4}{RGB}{229,241,212}  
\definecolor{color5}{RGB}{231,225,236}  
\definecolor{color6}{RGB}{217,238,238}  

\AtBeginDocument{%
  }

\copyrightyear{2026}
\acmYear{2026}
\setcctype{by}
\acmConference[WWW '26]{Proceedings of the ACM Web Conference 2026}{April 13--17, 2026}{Dubai, United Arab Emirates}
\acmBooktitle{Proceedings of the ACM Web Conference 2026 (WWW '26), April 13--17, 2026, Dubai, United Arab Emirates}
\acmPrice{}
\acmISBN{979-8-4007-2307-0/2026/04}
\acmDOI{10.1145/3774904.3792585}
\begin{document}

\title[LLM-centric Affective Visual Customization]{Towards LLM-centric Affective Visual Customization via Efficient and Precise Emotion Manipulating}

\author{Jiamin Luo}
\email{20204027003@stu.suda.edu.cn}
\affiliation{
  \institution{School of Computer Science and Technology, Soochow University}
  \city{Suzhou}
  \country{China}
}

\author{Xuqian Gu}
\email{xqgu0825@stu.suda.edu.cn}
\affiliation{
  \institution{School of Computer Science and Technology, Soochow University}
  \city{Suzhou}
  \country{China}
}

\author{Jingjing Wang}
\authornote{Corresponding author: Jingjing Wang}
\email{djingwang@suda.edu.cn}
\affiliation{
  \institution{School of Computer Science and Technology, Soochow University}
  \city{Suzhou}
  \country{China}
}

\author{Jiahong Lu}
\email{20255227109@stu.suda.edu.cn}
\affiliation{
  \institution{School of Computer Science and Technology, Soochow University}
  \city{Suzhou}
  \country{China}
}

\renewcommand{\shortauthors}{Jiamin Luo, Xuqian Gu, Jingjing Wang, \& Jiahong Lu}

\begin{abstract}
  Previous studies on visual customization primarily rely on the objective alignment between various control signals (e.g., language, layout and canny) and the edited images, which largely ignore the subjective emotional contents, and more importantly lack general-purpose foundation models for affective visual customization. With this in mind, this paper proposes an \textbf{L}LM-centric \textbf{A}ffective \textbf{V}isual \textbf{C}ustomization (L-AVC) task, which focuses on generating images within modifying their subjective emotions via Multimodal LLM. Further, this paper contends that how to make the model efficiently align emotion conversion in semantic (named inter-emotion semantic conversion) and how to precisely retain emotion-agnostic contents (named exter-emotion semantic retaining) are rather important and challenging in this L-AVC task. To this end, this paper proposes an \textbf{E}fficient and \textbf{P}recise \textbf{E}motion \textbf{M}anipulating (EPEM) approach for editing subjective emotions in images. Specifically, an \textbf{E}fficient \textbf{I}nter-emotion \textbf{C}onverting (EIC) module is tailored to make the LLM efficiently align emotion conversion in semantics before and after editing, followed by a \textbf{P}recise \textbf{E}xtra-emotion \textbf{R}etaining (PER) module to precisely retain the emotion-agnostic contents. Comprehensive experimental evaluations on our constructed L-AVC dataset demonstrate the great advantage of the proposed EPEM approach to the L-AVC task over several state-of-the-art baselines. This justifies the importance of emotion information for L-AVC and the effectiveness of EPEM in efficiently and precisely manipulating such information.
\end{abstract}



\begin{teaserfigure}
\begin{center}
    \includegraphics[width=\textwidth, scale=0.27]{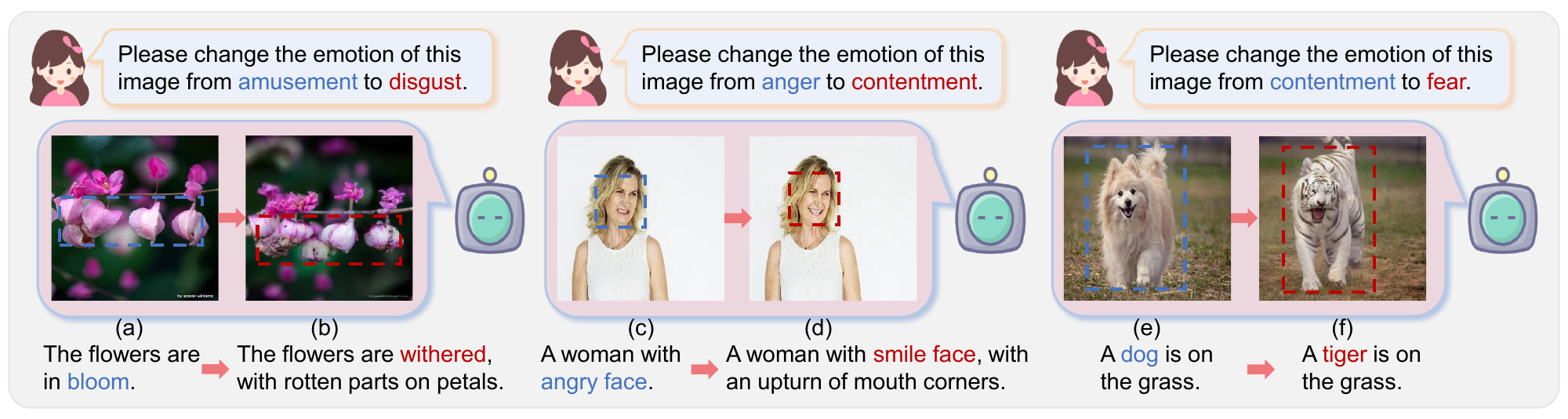}
\setlength{\abovecaptionskip}{-3 ex}
\setlength{\belowcaptionskip}{0 ex}
\caption{Three samples to illustrate the L-AVC task. The red boxes indicate differences from the original images.}
\label{fig:intro}
\end{center}
\end{teaserfigure} 


\maketitle

\section{Introduction}
\label{sec:intro}
Visual customization routinely focuses on employing various control conditions (e.g., natural language~\cite{Prompt-to-Prompt}, layout~\cite{LayoutLLM-T2I} and canny~\cite{ControlNet}) to edit objects in images that adhere to these conditions, which has been widely integrated into many visual design tools, such as Meitu and Snapseed. Most existing studies~\cite{InstructPix2Pix,DiffEdit} on visual customization leverage different generative models (e.g., GANs~\cite{gan}, VAEs~\cite{vae} and diffusion models~\cite{DDPM}) to generate the edited images, while a small amount of recent studies~\cite{MGIE,SmartEdit} have explored the use of Multimodal Large Language Models (MLLMs) to assist diffusion models aligning conditions and images. Despite these efforts achieve great progress, they focus on editing objective concepts (e.g., objects like \emph{a cat} or \emph{a house}) in images, and are limited in manipulating subjective emotions (e.g., \emph{excitement} or \emph{disgust}) inherent in images. Particularly, a few studies~\cite{EmoGen,EmoEdit} have focused on generating emotion perception images, while they are not chat-paradigm, making it difficult to understand editing instructions and adapt to the requirements of user interaction in the era of Large Language Models (LLMs).

Building on these observations, this paper proposes a novel \textbf{L}LM-centric \textbf{A}ffective \textbf{V}isual \textbf{C}ustomization (L-AVC) task, aiming to manipulate subjective emotions within images, which can significantly enhance emotional resonance of users and effectively inhibit the generation of harmful, biased and other unethical images in the current AIGC era. Specifically, the L-AVC task generates edited images that match emotion requirements via given instructions and original images. However, how to edit the inherently subjective emotions? Psychological studies~\cite{emotion-psychopathology1} have demonstrated that specific visual elements are often responsible for evoking emotions. Thus, this paper serves subjective emotions as a bridge between visual elements (i.e., \emph{facial}, \emph{action}, \emph{scene}, \emph{object} and \emph{color\&brightness} as reported by~\citet{EmoSet}) and the emotions they evoke. Taking Figure~\ref{fig:intro} (a) as an example, given the instruction ``\emph{Please change the emotion of this image from anger to amusement}'', the model requires to generate an image where manipulating the \emph{angry face} (\emph{anger}) to \emph{happy face} (\emph{amusement}) and retain other visual elements unchanged. In light of this, we explore two kinds of challenges for addressing the L-AVC task.

For one thing, how to make the model efficiently align emotion conversion in semantic is challenging, namely inter-emotion semantic conversion challenge. Emotions are inherently abstract, whereas images are concrete, creating a gap between the two. Traditional diffusion-based visual customization methods are difficult to understand how to edit the image to achieve emotion conversion, thus we introduce MLLMs to assist diffusion in understanding the editing instructions. However, emotions in the current MLLMs training data are consistent between images and captions. For instance in Figure~\ref{fig:intro} (b), we obtain the caption ``\emph{A woman with angry face}'' corresponding \emph{anger} emotion, but we need MLLMs to generate the caption ``\emph{A woman with smile face, with an upturn of mouth corners}'' that matches \emph{contentment} emotion according to the editing instruction, where the semantic consistency between image and caption induces the emotion conversion. To achieve this goal, we need a large amount of these aligned corpora to make the MLLM understand how to edit to be consistent with emotion conversion, while constructing such parallel corpora is difficult and expensive. Fortunately, some studies~\cite{Sparsityy1,Sparsity2} recently attempt to leverage model editing~\cite{knowledge-edit, mend} to achieve efficient alignment on low-resource corpus. Inspired by these studies, this paper explores the use of low-cost model editing to make the MLLM fully understand the semantic conversion between different emotions, and achieve efficient inter-emotion alignment. Therefore, this paper believes that a well-designed approach should consider employing model editing to efficiently align emotion conversion in semantic, rather than relying on the expensive alignment with large-scale data.      

For another, how to precisely retain emotion-agnostic semantics is challenging, namely exter-emotion semantic retaining challenge. Emotions are complex and are always triggered by various visual stimuli as reported by~\citet{emotion-psychopathology2}. Taking Figure~\ref{fig:intro} (a) as an example, changing the \emph{angry face} to the \emph{smiling face}, the emotion changes from \emph{anger} to \emph{amusement}. If we adjust the color and brightness to black and white at the same time, the emotion will shift from \emph{anger} to \emph{sadness}, which fails to meet the requirement of the \emph{amusement} emotion. Under this circumstance, we need to retain the original emotion-agnostic contents of the images as much as possible, aiming to avoid editing interference with the desired emotion. Therefore, this paper believes that a better-designed approach should consider retaining emotion-agnostic contents during the editing process to precisely meet the requirements of emotions.

To tackle the above challenges, this paper proposes an \textbf{E}fficient and \textbf{P}recise \textbf{E}motion \textbf{M}anipulating (EPEM) approach, consisting of two novel modules to L-AVC: (1) \textbf{E}fficient \textbf{I}nter-emotion \textbf{C}onverting (EIC) module introduces the model editing mechanism~\cite{multi-edit} to modifying MLP layers in LLMs, which efficiently aligns the emotion conversion between original and editing semantics to address the inter-emotion semantic conversion challenge. (2) \textbf{P}recise \textbf{E}xter-emotion \textbf{R}etaining (PER) module innovatively designs an \textbf{E}motion \textbf{A}ttention \textbf{I}nteraction (EAI) block to learn the interaction between the MLLM and diffusion model, which precisely retain the emotion-agnostic contents inside images to address the extra-emotion precise retaining challenge. Furthermore, a high-quality L-AVC dataset and a series of emotional metrics are tailored to evaluate the effectiveness of EPEM. Detailed evaluations demonstrate that EPEM could significantly outperform a bunch of state-of-the-art baselines, justifying the effectiveness of EPEM in editing subjective emotions inside images across various visual elements.

\begin{figure*}
    \centering
    \includegraphics[width=\textwidth, scale=0.32]{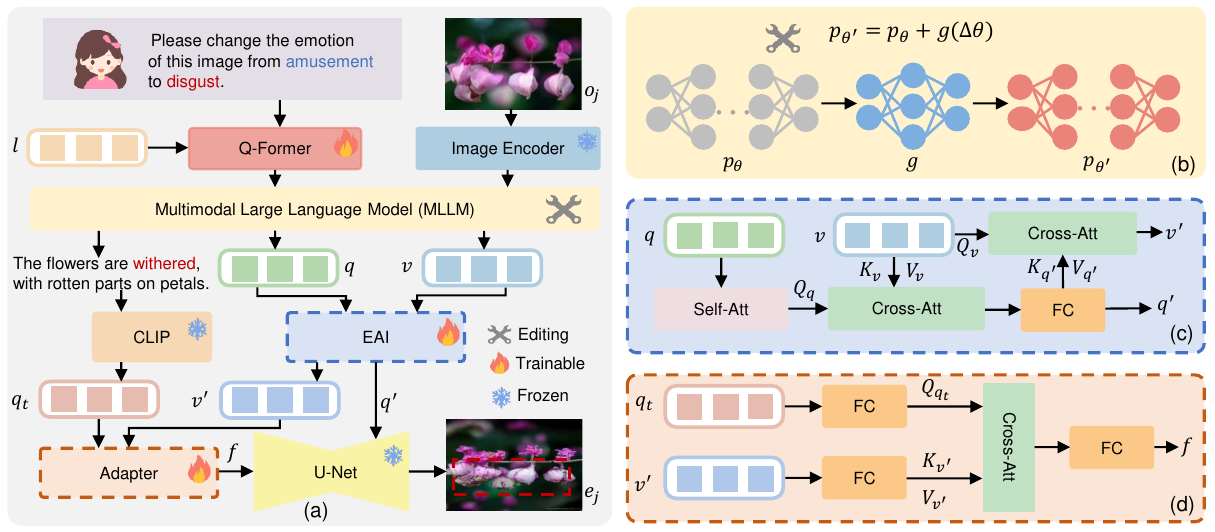}
    \setlength{\abovecaptionskip}{-3 ex}
    \setlength{\belowcaptionskip}{-3 ex}
    \caption{The overall architecture of our proposed EPEM approach (a). Wherein (b) shows the process of model editing in the EIC module (see Section~\ref{sec:EIC}), while (c) and (d) show the process of interaction between MLLM and diffusion in the PER module (see Section~\ref{sec:PER}), and FC represents the fully-connected layer.}
    \label{fig:model}
\end{figure*}
\section{Related Work}
\textbf{$\bullet$ Visual Customization.} The rise of diffusion models has ignited a surge in the field of image generation~\cite{HR-LDM,DM-GAN}. Building upon this, a more challenging research visual customization has emerged, which focuses on modifying desired elements while maintaining semantic-unrelated content. Previous studies can be broadly categorized into two main categories. Local editing focuses on modifying specific areas with an image, and can be subdivided into four types. \textbf{(1)} Object editing involves adding, removing, or replacing designated objects, including Prompt-to-Prompt~\cite{Prompt-to-Prompt}, Null-text Inversion~\cite{Null-text}, InstructPix2Pix~\cite{InstructPix2Pix}, etc. \textbf{(2)} Attribute editing focuses on enhancing, altering, or diminishing intrinsic object attributes, including MasaCtrl~\cite{MasaCtrl}, Imagic~\cite{Imagic}, etc. \textbf{(3)} Spatial transformation involves changing the spatial property of objects, including DragDiffusion~\cite{DragDiffusion}, SmartEdit~\cite{SmartEdit}, DragonDiffusion~\cite{DragonDiffusion}, etc. \textbf{(4)} Inpainting aims to fill the interesting area in the original image with contextually coherent content, including BLD~\cite{BLD} and HR-BLD~\cite{HR-BLD}. Global editing focuses on modifying the overall semantic content of the original image, and can be categorized into two types. \textbf{(1)} Style editing involves changing the style of the original image to that of another, including Cross-Image Attention~\cite{Cross-Image} and Style-Aligned Editing~\cite{Style}. \textbf{(2)} Image translation focuses on transferring an image from a source domain to a target domain, including ControlNet~\cite{ControlNet}, UniControl~\cite{UniControl}, etc. Besides, few studies~\cite{emotion-acc,EmoGen,EmoEdit} gradually note the generation of emotion perception images, while they are not chat-paradigm, making it difficult to understand instruction and adapt to interaction needs. In summary, although all the above studies have achieved significant success in visual customization, they are not able to achieve efficient and precise affective customization. Besides, they are not chat-paradigm, making it difficult to understand editing instructions and adapt to the requirements of user interaction in the era of LLMs. Unlike prior studies, this paper proposes a new LLM-centric affective visual customization (L-AVC) task, aiming to manipulate subjective emotions inside images.

\noindent\textbf{$\bullet$ LLM-assisted Diffusion Models.} Recently, diffusion models~\cite{DDPM,Score-DM} have advanced the state-of-the-art across a range of vision generation tasks,like text-to-image generation. Given the remarkable semantic understanding capabilities of LLMs, some studies have explored leveraging their strengths to enhance the text-conditioned generation performance of diffusion models. Specifically, \citet{LayoutLLM-T2I} utilize ChatGPT~\cite{chatgpt} to generate image layouts, and employ a diffusion model conditioned on both textual prompts and layouts to synthesize images. \citet{SUR-adapter} transfer knowledge from LLMs to diffusion models via an adapter, enabling them to understand and reason with concise language. Furthermore, recognizing the need for alignment between image and text, several studies leverage MLLMs to assist diffusion models in generation tasks. Specifically, GILL~\cite{GILL} learns to process images with LLMs and generate coherent images. SEED~\cite{SEED} introduces an image tokenizer that allows LLMs to process and generate images and text concurrently. MGIE~\cite{MGIE} integrates MLLMs with editing instructions, learning to derive expressive instructions and providing explicit guidance for visual customization. SmartEdit~\cite{SmartEdit} jointly trains LLaVA~\cite{llava} and diffusion models for instruction-based visual customization. Particularly, few studies explore the generation of emotions in images. For instance, \citet{EmoEdit} leverage GPT-4V~\cite{gpt4v} offline to generate editing instructions via emotion factor trees to synthesize images. Different from all the above studies, this paper introduces model editing to assist the MLLM in understanding emotion conversion for addressing the inter-emotion semantic conversion challenge, and then designs an emotional interaction block to ensure the invariance of emotion-agnostic contents for tackling the extra-emotion semantic retaining challenge.

\section{Approach}
\label{sec:approach}
In this section, we formulate the L-AVC task as follows. Given a set of original images $O=\{o_1, \cdots, o_i, \cdots, o_n\}$ with original image captions $C=\{c_1, \cdots, c_i, \cdots, c_n\}$ (e.g., \emph{The flowers are in bloom}), emotion labels $Y=\{y_1, \cdots, y_i, \cdots, y_n \}$ (e.g., \emph{amusement}) and editing instructions $X=\{x_1, \cdots, x_i, \cdots, x_n \}$ (e.g., \emph{Please change the emotion of this image from amusement to disgust.}), where $n$ represents the number of images, the goal of the L-AVC task is to generate corresponding edited images set $E=\{e_1, \cdots, e_i, \cdots, e_n\}$ that matches the edited image captions $\hat{C}=\{\hat{c}_1, \cdots, \hat{c}_i, \cdots, \hat{c}_n \}$ (e.g., \emph{The flowers are withered, with rotten parts on petals.}) and emotions $\hat{Y}=\{\hat{y}_1, \cdots, \hat{y}_i, \cdots, \hat{y}_n \}$ (e.g., \emph{disgust}), while retaining the emotion-agnostic contents of $O$.

In this paper, we propose an \textbf{E}fficient and \textbf{P}recise \textbf{E}motion \textbf{M}anipulating (EPEM) approach to address the L-AVC task as illustrated in Figure~\ref{fig:model} (a), involving inter-emotion semantic conversion challenge and exter-emotion semantic retaining challenge. To tackle these challenges, we design an \textbf{E}fficient \textbf{I}nter-emotion \textbf{C}onverting (EIC) module and a \textbf{P}recise \textbf{E}xter-emotion \textbf{R}etaining (PER) module. Prior to delving into the intricacies of the core components within EPEM, we provide an overview of the diffusion models. 

\subsection{Preliminary of Diffusion Models}
In this study, we use the open-source Stable Diffusion (SD)~\cite{SD} as our backbone, owing to its state-of-the-art generative capabilities. The core mechanism of SD involves systematically perturbing the structure of data distributions (named forward diffusion) and recovering the structure via denoising (named reverse diffusion). Specifically, given an image $o_i$, SD first obtain the noising image vector $\{z_t\}_{t=0}^T=\mathcal{E}(o_i)$ via the image encoder $\mathcal{E}$, where $T$ represents the time step. Then, the forward diffusion process from $z_t$ to $z_{t-1}$ in the latent space can be modeled by: $q(z_t | z_{t-1}) = \mathcal{N}(z_t; \alpha_{t}z_{t-1}, (1-\alpha_{t}^2)\mathbf{I})$, where $q(z_t | z_{t-1})$ denotes the posterior distribution, and $\alpha_{t} \in (0,1)$ is constant hyper-parameters. $\mathcal{N}$ represents the normal distribution and $\mathbf{I}$ represents the identity covariance matrix.

Then, at the reverse diffusion process leverage a UNet $\epsilon_\theta$ trained to predict the noise $\epsilon$ added from $z_0$ to $z_t$, given the image condition $v$ and text instruction $h$, i.e., $\epsilon = \epsilon_\theta(t, \textrm{concat}[z_t, v], h)$, where $v=\mathcal{E}(o_i)$, which is incorporated by directly concatenating $v$ and $z_t$. $h$ is the embedding of text instructions encoded by CLIP~\cite{clip} in the L-AVC task. Therefore, the denoised $z_{t-1}$ of $z_t$ at each time step $t$ can be denoted as: $z_{t-1} = z_t - \epsilon_\theta(t, \textrm{concat}[z_t, v], h)$. During the inference stage, $z_0$ is obtained with $T$ iterative denoising steps given the timestep $t$, latent noising image vector $z_t \in \mathcal{N}(\mathbf{0},\mathbf{I})$, image condition $v$ and conditioning text embedding $q_t$. Then, the image decoder maps $z_0$ to the original image $o_i$.

\subsection{Efficient Inter-emotion Converting Module}
\label{sec:EIC}
In this study, we take advantage of model editing mechanism~\cite{mend} and propose an \textbf{E}fficient \textbf{I}nter-emotion \textbf{C}onverting (EIC) module to make LLM understand the emotion semantic conversion from original emotion $y_i$ to editing emotion $\hat{y}_i$ according to the instruction $x_i$. In contrast to modifying factual knowledge~\cite{knowledge-edit,mend}, we focus on learning the semantic alignment between original and editing emotions via updating parameters in the LLM. As shown in Figure~\ref{fig:model} (b), we employ a hyper-network $g$ to modify the parameters $\theta$ of the LLM model, resulting in updated parameters $\theta'$, formulated as: 
\begin{equation}
\label{eq:editing}
 p_{\theta'} = p_{\theta} + g(\Delta \theta)  
\end{equation}
where $p_{\theta}$ is the model with trainable parameters $\theta$, and $\Delta \theta$ is a shift predicted by the hyper-network $g$.

Specifically, given a dataset of editing inputs $(O, C, Y, X) \in \mathcal{D}$, we aim to leverage the LLM to generate the edited image captions $\hat{C}$ and emotions $\hat{Y}$, that is the model $p_{\theta}$ mapping these inputs in $\mathcal{D}$ and a set of parameters $\theta$ to obtain the image captions $\hat{C}$ within emotions $\hat{Y}$. To achieve the goal, we leverage the model $p_{\theta}$ with the editing dataset $\mathcal{D}$ to produce a collection of model editor hyper-network $g$, which edit the weights matrices $\mathbf{W}$ of the model $p_{\theta}$ to learn the updated model $p_{\theta'}$ with new parameters $\theta'$.

Since editing attention layers in the LLM brings performance reduction~\cite{mend}, we select the weights matrix $\mathbf{W}=\{\mathbf{W}_1, \cdots, \mathbf{W}_N\}$ of MLP layers in the LLM that we would like to make editable before training, such as the weight matrices in the last $N$ MLP layers. At each training step, we sample the edit sample $(o, c, y, x)$ from the editing dataset $\mathcal{D}$ to compute the raw gradient for each weight matrix $\mathbf{W}_i \in \mathbf{W}$. Then, we compute the parameter update $\Delta \mathbf{W}$ for each layers $i$ as:
\begin{equation}
\label{eq:update}
 \Delta \mathbf{W} = \mathbf{W}_i - \alpha \nabla_{{\rm W}_i}
\end{equation}
where $\nabla_{{\rm W}_i}$ denotes the edited weight matrix $\mathbf{W}_i$ by the model editor hyper-network $g$, where each $g$ transforms gradient for a specific layer $i$ into a parameter shift $\Delta \theta$ for that layer to capture the alignment of emotions. $\alpha$ is a learned per-layer scalar step size.

\subsection{Precise Exter-emotion Retaining Module}
\label{sec:PER}
In this study, we take advantage of the attention mechanism in visual customization~\cite{Prompt-to-Prompt} and propose a \textbf{P}recise \textbf{E}xter-emotion \textbf{R}etaining (PER) module to precisely retain emotion-agnostic contents inside original images $o$, aiming to meet the requirements of edited emotions $\hat{y}$ in instructions $x$. As illustrated in Figure~\ref{fig:model}, we aim to connect the MLLM and the SD model through a tailored \textbf{E}motional \textbf{A}ttention \textbf{I}nteraction (EAI) block and an adapter.

Specifically, the EAI block consists of a self-attention block, two cross-attention blocks and an MLP layer as shown in Figure~\ref{fig:model} (c). The inputs to EAI are the output $q$ from the Q-Former $\mathcal{Q}$ and the output $v$ from the image encoder $\mathcal{E}$. Particularly, $q$ originates from the Q-Former in BLIP2, which addresses the discrepancy between the feature spaces of the BLIP2~\cite{blip2} and SD model, denoted as $q=\mathcal{Q}(v, x, l)$, where $l$ are learnable queries. $v$ is encoded by $\mathcal{E}$ to interacts with $q$, denoted as $v=\mathcal{E}(o)$.

For the interaction between $q$ and $v$, the process begins by applying a self-attention block to $q$. Then, $q$ serves as the query $\mathbf{Q}_q$ in a cross-attention block to interact with $v$, which functions as both key $\mathbf{K}_v$ and value $\mathbf{V}_v$. This interaction generates $q'$ through a point-wise MLP, formulated as follows:
\begin{equation}
    \label{eq: eai_q}
    q' = \textrm{MLP}(\bm{{\rm V}_{v}} \cdot \textrm{softmax}(\bm{{\rm Q}_{q}}^\top \bm{{\rm K}_v}))
\end{equation}
After the creation of $q'$, it serves as both key $\mathbf{K}_{q'}$ and value $\mathbf{V}_{q'}$ in another cross-attention, with $v$ acts as the query $\mathbf{Q}_v$, resulting in the generation of $v'$, formulated as follows:
\begin{equation}
    \label{eq: eai_v}
    v' = \bm{{\rm V}_{q'}} \cdot \textrm{softmax}(\bm{{\rm Q}_{v}}^\top \bm{{\rm K}_{q'}})
\end{equation}

To make use of the impressive image generation capabilities of SD model and further reduce the computational cost, we leverage an adapter to tune the frozen SD model. Specifically, the adapter consists of three MLP layers and a cross-attention block as shown in Figure~\ref{fig:model} (d). The inputs to the adapter are the output $q_t$ from the CLIP model~\cite{clip} and the output $v'$ from the EAI block. Particularly, the CLIP model encodes the edited instructions $x$ and image captions $\hat{c}$ as the conditions, denoted as $q_t=\textrm{CLIP}(x,c)$. Two MLP layers are leveraged to transform $q_t$ and $v'$, and $q_t$ serves as query $\mathbf{Q}_{q_t}$ while $v'$ serves as key $\mathbf{K}_{v'}$ and value $\mathbf{V}_{v'}$ in the cross-attention block. Then, the output of the adapter $f$ through the third MLP layer, formulated as follows:
\begin{equation}
    \label{eq: eai_v}
    f = \textrm{MLP}(\bm{{\rm V}_{v'}} \cdot \textrm{softmax}(\bm{{\rm Q}_{q_t}}^\top \bm{{\rm K}_{v'}}))
\end{equation}

\subsection{Optimization for EPEM approach}
To address the L-AVC task, we employ a training strategy to optimize our EPEM approach. 

For EIC module, we introduce a loss $\mathcal{L}_{\rm EIC}$ formulated as:
\begin{equation}
    \label{eq:loss_EIC}
    \mathcal{L}_{\rm EIC} = -\log p_{{\theta'}_{\rm W}} (\hat{c}, \hat{y}|o, c, y, x) - \lambda_1 \cdot \sum\nolimits_{i}\hat{y}_{i}\log p(\hat{c}_{i})
\end{equation}
The first term is used to train the parameters $\theta'$ within weights $\rm W$ of the edited model, which measures the edit success through back-propagation. ${\theta'}_{\rm W}$ denoting the model parameters $\theta'$ of the unedited weights weights $\rm W$. The second term is designed to ensure semantic alignment between original and editing emotions, which measures the difference between the generated editing captions $\hat{c}$ and desired editing emotions $\hat{y}$, where $p(\hat{c}_{i})$ represents the prediction probability of emotions for the image caption $\hat{c}_{i}$, and $\lambda_1$ is loss coefficient to balance two terms.

For PER module, we introduce a loss $\mathcal{L}_{\rm PER}$ formulated as: 
\begin{equation}
    \label{eq:loss_PER}
    \mathcal{L}_{\rm PER} = \mathbb{E}\left\| \epsilon-\epsilon_{\theta}\left(t, \textrm{concat}(z_{t}, v), h\right) \right\|_{2}^{2} 
    + \frac{\lambda_2}{M} \cdot \sum_{j=1}^{M}\left\|o_j - e_j \right\|^{2}
\end{equation}
The first term is used to ensure that the pre-trained SD model retains sufficient denoising ability for new images $e$ during fine-tuning, where $h$ represents the conditions including $f$ and $q'$. Besides, to ensure stable training of the added adapter and mitigate its adverse impact on the pre-trained SD model during the early stage, we follow previous works~\cite{lora,ControlNet} by initializing all elements of the matrices in parameter $\theta$ to 0. The second term is designed to retain emotion-agnostic contents of the image $o$, which measures the similarity between the original image $o$ and the editing image $e$ on the pixel level, where $o_j$ and $e_j$ represent the original and editing pixel values, and $M$ represents the total number of pixels. $\lambda_2$ is another loss coefficient to balance two terms in the PER module. Therefore, the final loss for training denotes as $\mathcal{L} = \mathcal{L}_{\rm EIC} + \mathcal{L}_{\rm PER}$.

\begin{figure}[t]
  \centering
  \includegraphics[width=\columnwidth]{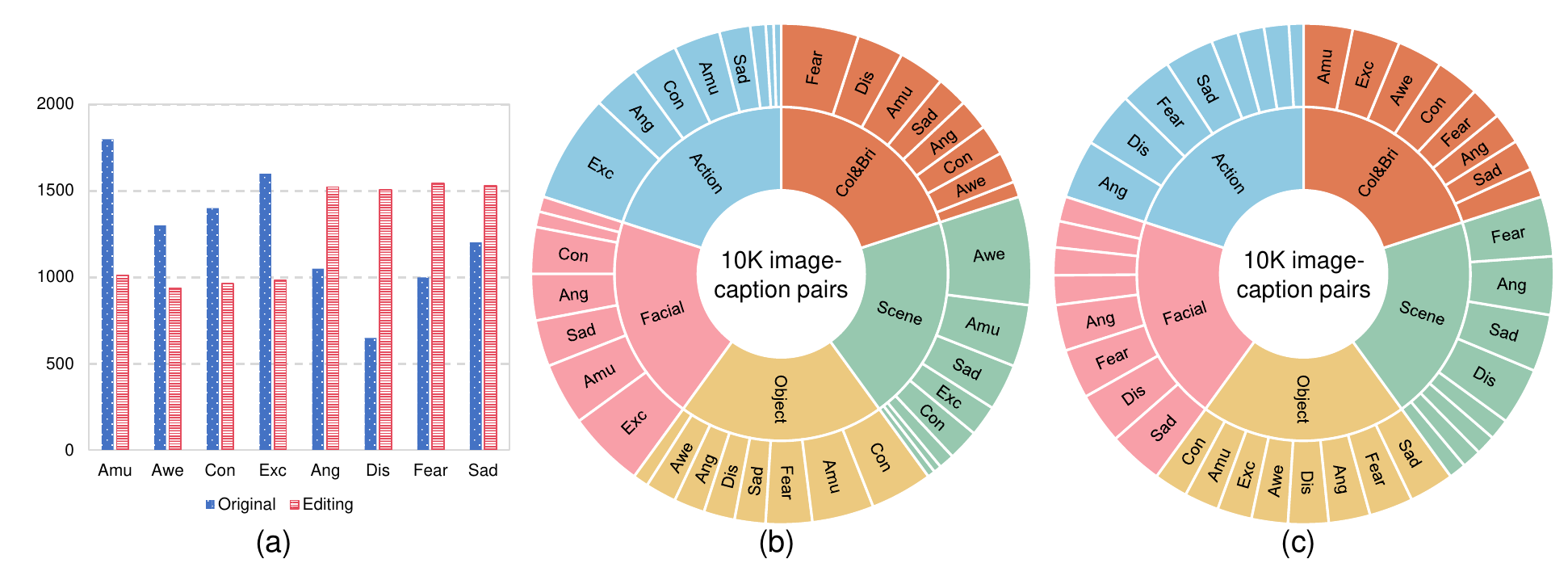}
  \setlength{\abovecaptionskip}{-2 ex}
  \setlength{\belowcaptionskip}{-2 ex}
  \caption{A histogram (a) to illustrate statistics of our L-AVC dataset before and after editing, and two fan charts to show the emotional distribution of different visual elements, with (b) and (c) representing pre-editing and post-editing.}
  \label{fig:dataset}
\end{figure}

\section{Experimental Settings}

\setlength{\tabcolsep}{1.8pt}
\begin{table*}[]
\renewcommand{\arraystretch}{0.9}
\addtolength{\tabcolsep}{6pt}
\begin{center}
\setlength{\abovecaptionskip}{-0.1 ex}
\setlength{\belowcaptionskip}{-0.5 ex}
\caption{Comparison of several visual customization models and EPEM on L-AVC dataset. The $\downarrow$ beside FID and LPIPS indicates the lower metric means better performance. Bold and underlined indicate the highest and second-highest performance.}
\setlength{\belowcaptionskip}{-5 ex}
\label{tab:main_results}
\resizebox{\linewidth}{!}{
    \begin{tabular}{c|cccc|c|ccc|c}
\toprule[1.2pt]
\multirow{2}{*}{Approach} & \multicolumn{4}{c|}{Consistency}                          & Fidelity & \multicolumn{3}{c|}{Accuracy}          & Efficiency \\ \cline{2-10} 
                                   & FID$\downarrow$   & LPIPS$\downarrow$ & SSIM  & CLIP-I & CLIP-T   & M-Eval  & G-Eval  & H-Eval & Time        \\ \hline
ControlNet                & 0.140          & 0.596          & 29.38          & 69.24           & 20.64             & 38.61          & 42.4          & 58.4           & 8.9                 \\
Prompt-to-Prompt                       & 0.131          & 0.558          & 45.64          & 73.27           & 21.57             & 45.22          & 53.4          & 65.0           & 7.6                 \\
InstructPix2Pix                   & 0.125          & 0.508          & 42.71          & 72.76           & 22.56             & 47.42          & 56.6          & 66.2           & 7.2                 \\
SDEdit                    & 0.115          & 0.441          & 47.67          & 71.08           & 22.45             & 51.08          & 60.2          & 68.8           & 7.1                 \\
DiffEdit                  & 0.118          & 0.426          & 49.68          & 72.35           & 21.19             & 52.45          & 63.6          & 69.0           & 7.3                 \\
MGIE                      & 0.099          & {\ul 0.390}    & {\ul 53.56}    & {\ul 79.18}     & {\ul 24.92}       & {\ul 52.82}    & {\ul 68.0}    & {\ul 73.0}     & 9.8                 \\
SmartEdit                 & {\ul 0.098}    & 0.397          & 52.08          & 78.89           & 23.31             & 51.00          & 67.8          & 71.8           & 10.3                \\ \hline
\textbf{EPEM}                       & \textbf{0.068} & \textbf{0.339} & \textbf{58.29} & \textbf{82.70}  & \textbf{27.96}    & \textbf{59.97} & \textbf{75.8} & \textbf{80.2}  & 9.6                \\
\rowcolor{color1} w/o EIC                & 0.107          & 0.435          & 48.31          & 73.26           & 21.93             & 51.75          & 60.4          & 65.4           & 7.5                 \\
\rowcolor{color2} w/o EAI                   & 0.102          & 0.405          & 51.64          & 77.74           & 23.15             & 53.74          & 64.6          & 69.2           & 9.1                 \\ 
\bottomrule[1.2pt]
\end{tabular}
}
\end{center}
\end{table*}

\subsection{Dataset Construction}
To evaluate the effectiveness of our EPEM approach for the L-AVC task, we construct an L-AVC dataset. Inspired by~\citet{lisa} that a minimal amount of reasoning segmentation data can effectively activate the reasoning capacity of MLLMs, we randomly select 2K images from the EmoSet dataset~\cite{EmoSet}, encompassing five visual elements: facial, action, object, scene, col\&bri (short for color and brightness), to form a total of 10K images. These images are split into training and test sets in an 8:2 ratio. Detailed statistics of the L-AVC dataset are shown in Figure \ref{fig:dataset}. 

The construction of this L-AVC dataset can be specifically divided into two parts. 
\textbf{(1) Editing Instruction Construction.} Specifically, due to the lack of image captions in the Emoset, we first utilize GPT-4V~\cite{gpt4v} to generate captions for each image. Subsequently, we employ five annotators to annotate the edited emotion of each generated caption. The detailed annotations include the specific visual elements being edited in the caption, post-edited emotion and the editing instruction. For example in Figure~\ref{fig:intro} (b), given an image with \emph{anger} emotion and caption ``\emph{A woman with angry face}'' via GPT-4V, we need to edit the visual element \emph{angry face} to \emph{smile face}, achieving the emotion conversion from \emph{anger} to \emph{contentment}. Thus, we design the instruction ``\emph{Please change the emotion of this image from anger to contentment}''. Finally, we assign another expert to verify the alignment between the edited captions and emotions, and check whether the emotions of editing visual elements are consistent with the corresponding edited emotions. 
\textbf{(2) Editing Image Generation.} Given the labor-intensive and time-consuming cost of manually editing each image, we first leverage InstructPix2Pix~\cite{InstructPix2Pix} followed by \citet{SmartEdit} to generate the synthetic edited images and then adjust them to match the edited emotions. Specifically, we use the conditions of the original images and edited captions to generate the edited images. Then, we employ five annotators to manually filter the unmatched images, and again adjust conditions to generate new synthetic edited images. Particularly, to better evaluate the effectiveness of our EPEM approach, we meticulously adjust the unreasonable generated images in the test set. Finally, we assign another expert to check the alignment between the edited emotions and images, as well as between the edited captions and images. After verification of editing instruction construction and image generation, we integrate the 10K editing instructions, original and edited images, and emotions before and after editing to compile the rest part of the L-AVC dataset. Owing to the subjectivity of emotions, it is difficult to calculate the \emph{Kappa} value, thus we only ensure the rationality of edited emotion during the final verification.

\subsection{Baselines}
We choose several advanced baselines in visual customization benchmark against our EPEM approach, described as follows. \textbf{ControlNet}~\cite{ControlNet} is an end-to-end network that leverages the large pre-trained model as a strong backbone to learn diverse conditional controls. \textbf{Prompt-to-Prompt}~\cite{Prompt-to-Prompt} provides an intuitive visual customization interface by editing only the textual prompt, without the need for model training, fine-tuning or optimization. \textbf{InstructPix2Pix}~\cite{InstructPix2Pix} involves two steps: generating an visual customization dataset and training a diffusion model on that dataset. \textbf{SDEdit}~\cite{SDEdit} relies on a generative prior from the diffusion model and synthesizes realistic images by iterative denoising via a stochastic differential equation. \textbf{DiffEdit}~\cite{DiffEdit} utilizes text-conditioned diffusion models for semantic visual customization. \textbf{MGIE}~\cite{MGIE} leverages the diffusion model to jointly train and perform visual customization with latent imagination via an edit head in an end-to-end manner. \textbf{SmartEdit}~\cite{SmartEdit} jointly optimizes the MLLM and diffusion model, and leverages the powerful understanding and reasoning abilities of MLLM to facilitate instruction-based editing. Since these released models are trained on specific datasets, they would inevitably perform poorly if directly evaluated on our L-AVC dataset. To ensure a fair and thorough comparison, we fine-tune these models on the same training set used by our EPEM approach, and then evaluate the fine-tuned models on the L-AVC dataset.

\subsection{Implement Details}
In our experiments, we fine-tune all the baselines on our L-AVC dataset according to their open-source codes. Besides, the hyper-parameters of these baselines reported by their public papers are still adopting the same setting. For the training of our EPEM approach, we choose the BLIP2 with the OPT (2.7B)~\cite{opt} as the MLLM to be edited in the EIC module, and the frozen Stable Diffusion-v1.5 to generate edited images in the PER module. Specifically, we edit the MLP weights in the last 3 transformers blocks following~\citet{multi-edit}, since editing MLP layers generally provides better editing performance than editing attention layers. We adopt AdamW with the learning rate 1e-6 and weight decay 1e-4 to optimize EPEM. Besides, we set the batch size, epochs, $\lambda_1$ and $\lambda_2$ to be 1, 3, 0.5, 0.3. Note that the edited MLP layer, Q-Former, EAI block and adapter are fully optimized, while CLIP, Stable Diffusion-v1.5, the image encoder and the rest transformer blocks in BLIP2 are frozen during the training process. All the experiments are conducted in PyTorch on 1 A100 GPU with 40GB memory. 

\begin{table*}[]
\begin{center}
\setlength{\abovecaptionskip}{-0.1 ex}
\setlength{\belowcaptionskip}{-0.8 ex}
\caption{Comparison of several visual customization models and our EPEM approach on our L-AVC dataset to evaluate the emotion editing performance of each visual element, where Col\&Bri is short for the visual element color and brightness.}
\setlength{\belowcaptionskip}{-5 ex}
\label{tab:ablation_results}
\resizebox{\linewidth}{!}{
\begin{tabular}{c|ccc|ccc|ccc|ccc|ccc}
\toprule[1.2pt]
\multirow{2}{*}{Approach} & \multicolumn{3}{c|}{Action}              & \multicolumn{3}{c|}{Facial}              & \multicolumn{3}{c|}{Object}              & \multicolumn{3}{c|}{Scene}               & \multicolumn{3}{c}{Col\&Bri}                \\ \cline{2-16} 
                                   & FID$\downarrow$   & CLIP-I & H-Eval & FID$\downarrow$   & CLIP-I & H-Eval & FID$\downarrow$   & CLIP-I & H-Eval & FID$\downarrow$   & CLIP-I & H-Eval & FID$\downarrow$   & CLIP-I & H-Eval \\ \hline
ControlNet                & 0.163          & 63.59           & 39.4           & 0.162          & 63.19           & 52.4           & 0.129          & 71.93           & 63.2           & 0.124          & 70.91           & 65.4           & 0.124          & 76.58           & 71.6           \\
Prompt-to-Prompt                       & 0.140          & 67.45           & 45.2           & 0.157          & 68.66           & 57.8           & 0.123          & 75.13           & 71.6           & 0.114          & 75.06           & 73.6           & 0.122          & 80.05           & 76.8           \\
InstructPix2Pix                   & 0.134          & 68.78           & 46.8           & 0.139          & 67.41           & 56.4           & 0.124          & 73.80           & 72.6           & 0.113          & 74.63           & 76.8           & 0.116          & 79.16           & 78.4           \\
SDEdit                    & 0.121          & 65.37           & 54.0           & 0.128          & 65.11           & 61.6           & 0.113          & 73.65           & 73.2           & 0.104          & 72.42           & 76.0           & 0.109          & 78.83           & 79.2           \\
DiffEdit                  & 0.126          & 66.97           & 53.2           & 0.134          & 67.04           & 61.2           & 0.113          & 74.33           & 74.4           & 0.111          & 74.87           & 75.6           & 0.106          & 78.54           & 80.6           \\
MGIE                      & 0.108          & {\ul 75.05}     & {\ul 56.6}     & {\ul 0.111}    & {\ul 75.62}     & {\ul 69.6}     & 0.098          & 80.23           & 75.0           & {\ul 0.082}    & 80.49           & {\ul 79.0}     & 0.094          & {\ul 84.49}     & {\ul 84.8}     \\
SmartEdit                 & {\ul 0.103}    & 74.38           & 54.8           & 0.116          & 73.81           & 67.8           & {\ul 0.098}    & {\ul 81.41}     & {\ul 76.8}     & 0.083          & {\ul 80.91}     & 77.4           & {\ul 0.091}    & 83.95           & 82.2           \\ \hline
\textbf{EPEM}                       & \textbf{0.072} & \textbf{79.89}  & \textbf{70.6}  & \textbf{0.071} & \textbf{79.23}  & \textbf{76.0}  & \textbf{0.066} & \textbf{84.39}  & \textbf{89.4}  & \textbf{0.065} & \textbf{83.76}  & \textbf{88.8}  & \textbf{0.065} & \textbf{86.24}  & \textbf{91.2}  \\
\rowcolor{color1} w/o EIC                & 0.111          & 69.14           & 43.6           & 0.121          & 69.63           & 60.4           & 0.111          & 73.48           & 73.8           & 0.091          & 74.41           & 71.4           & 0.100          & 79.65           & 77.8           \\
\rowcolor{color2} w/o EAI                   & 0.107          & 74.52           & 50.4           & 0.110          & 73.41           & 64.8           & 0.104          & 79.29           & 76.6           & 0.098          & 79.09           & 74.2           & 0.093          & 82.41           & 80.0           \\ 
\bottomrule[1.2pt]
\end{tabular}
}
\end{center}
\end{table*}

\subsection{Evaluation Metrics}
To comprehensively evaluate the performance of various models on the L-AVC task, we leverage widely used editing metrics and introduce several emotion-specific ones. These evaluation metrics are summarized into four categories, described as follows.

$\bullet$ \textbf{Consistency of Editorial Content} is used to measure the consistency between original and edited images. To ensure that only the intended visual elements are edited while retraining emotion-agnostic contents, we adopt FID, LPIPS, SSIM and CLIP-I four metrics. Specifically, Frechet Inception Distance (FID)~\cite{FID} quantifies the distributional difference between original and edited images, with lower values indicating greater similarity to the original distribution. Learned Perceptual Image Patch Similarity (LPIPS)~\cite{LPIPS} measures perceptual similarity between images, where lower scores indicate better performance. Structural Similarity Index (SSIM)~\cite{PSNR-SSIM} evaluates image similarity by considering luminance, contrast, and structure, providing a better reflection of human-perceived image quality. CLIP-I~\cite{CLIP-score} assesses the cosine similarity of CLIP image embeddings between ground-truth and generated editing images.

$\bullet$ \textbf{Fidelity of Editorial Semantic} is used to measure how accurately edited images align with editing captions, and we utilize CLIP-T~\cite{CLIP-score} to compute the cosine similarity of CLIP embeddings between edited images and editing captions.

$\bullet$ \textbf{Accuracy of Editorial Emotion} is designed to measure the alignment between the intended emotions and the edited images. To achieve this goal, we design M-Eval, G-Eval and H-Eval three evaluation methods. Specifically, M-Eval uses the advanced MDAN~\cite{mdan} emotion analysis model to predict the emotion of each edited image, and compute the accuracy. G-Eval employs the state-of-the-art GPT-4V~\cite{gpt4v} to visually analyze and infer the emotion of edited images, with the designed instruction ``\emph{As an image emotion analysis assistant, please determine the emotion of this input image and choose the emotion from [amusement, awe, contentment, excitement, anger, disgust, sadness, fear]}''. H-Eval hires five annotators to manually label the emotion of each edited image. After collecting annotations from each annotator, we average the results to report the final score. Due to the expensive GPT-4V and human costs, we randomly select 100 samples for each visual element to evaluate. 

$\bullet$ \textbf{Efficiency of Editorial Image} is used to measure the time cost associated with edited images for each approach. Since computing the time cost of a single image is challenging, we compute the total editing time for all images in the test set and take the average as the final time result to report. Moreover, $t$-test\footnote{\url{https://docs.scipy.org/doc/scipy/reference/stats.html}} is used to evaluate the significance of the performance.  

\section{Results and Discussion}

\subsection{Experimental Results}
Table \ref{tab:main_results} shows the performance comparison of different approaches on the L-AVC dataset. From this table, we can see that: 
\textbf{(1)} Consistency of Editorial Content. The emotion editing image generated by our EPEM approach outperforms all baselines in terms of FID, LPIPs, SSIM and CLIP-I. For example, compared with the best results underlined on the FID, LPIPs, SSIM and CLIP-I four metrics, EPEM achieves the improvements of 3\%($p$-value<0.05), 5.1\%($p$-value<0.01), 4.73\%($p$-value<0.05) and 3.52\%($p$-value<0.05) respectively. These results indicate that the emotion editing image generated by our EPEM approach could retain the consistency of the original image contents.  
\textbf{(2)} Fidelity of Editorial Semantic. Our EPEM approach achieves comparable results in terms of in CLIP-T, indicating that EPEM could follow the semantics of editing instructions and generate emotion editing images, keeping the fidelity to the instructions.   
\textbf{(3)} Accuracy of Editorial Emotion. The emotion editing image generated by our EPEM approach outperforms all baselines in terms of M-Eval, G-Eval and H-Eval. For example, EPEM surpasses the state-of-the-art MGIE (an MLLM-assisted editing model) by 7.15\%, 7.8\% and 7.2\% in M-Eval, G-Eval and H-Eval respectively. Statistical significance tests show that these improvements are significant ($p$-value<0.01). This indicates that previous editing models even with the assistance of MLLMs could not effectively understand emotion conversion, which further justifies that our EPEM approach could precisely align the emotion semantics between original and edited emotions.  
\textbf{(4)} Efficiency of Editorial Image. Our EPEM approach can accomplish the emotion editing task in 10 seconds (9.6) for a single input on an NVIDIA A100 GPU (40GB), which is faster compared with MGIE (9.8) and SmartEdit (10.3) two MLLM-assisted editing models. This justifies the efficiency of our EPEM approach compared with other MLLM-assisted visual customization models.

\begin{figure}
\begin{center}
    \includegraphics[width=\linewidth]{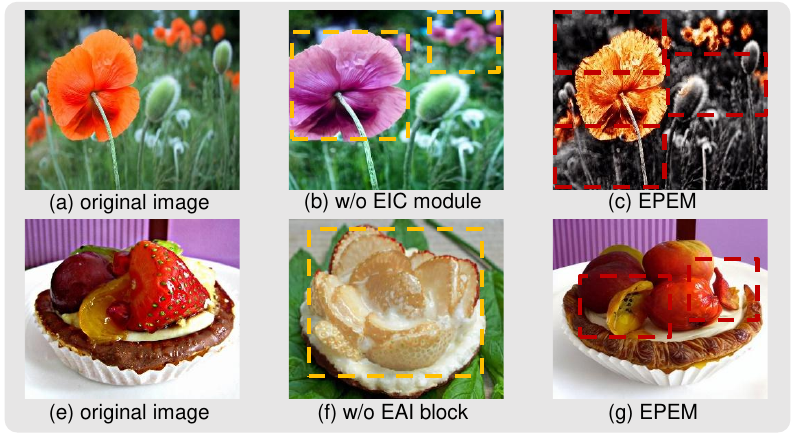}
    \setlength{\abovecaptionskip}{-2 ex}
    \setlength{\belowcaptionskip}{-2 ex}
\caption{Two samples to illustrate the precise inter-emotion conversion manipulating ((a), (b), (c)) and exter-emotion contents retraining ((d), (e), (f)) two challenges.}
\label{fig:analysis}
\end{center}
\end{figure} 

\subsection{Are Emotion Conversion Really Aligned?}
To validate the ability of our EPEM approach to understand the emotion conversion between original and edited emotions (i.e., inter-emotion semantic conversion challenge), we conduct quantitative and qualitative experiments on our L-AVC dataset.
\textbf{(1) Quantitative} results are shown in Table \ref{tab:ablation_results}. From this table, we can see that EPEM outperforms the best results underlined by an average improvement 9.84\% ($p$-value<0.01) in H-Eval on five visual elements, indicating that EPEM could enable the emotion conversion of specific semantics and visual elements compared to other baselines. This justifies EPEM could align the emotion semantic conversion according to the editing instruction. 
\textbf{(2) Qualitative} analysis is shown in Figure \ref{fig:analysis}. From this figure, we can intuitively observe that EPEM without EIC module changes the color of the flower (b), but the edited emotion (\emph{amusement}) is still the same as the original image (a), and our EPEM approach changes the \emph{amusement} emotion to \emph{sadness} by changing the color and brightness of the image (c). This demonstrates that EPEM can understand the alignment of emotion semantic conversion through the EIC module.

\begin{figure*}
\begin{center}
    \includegraphics[width=\textwidth, scale=0.2]{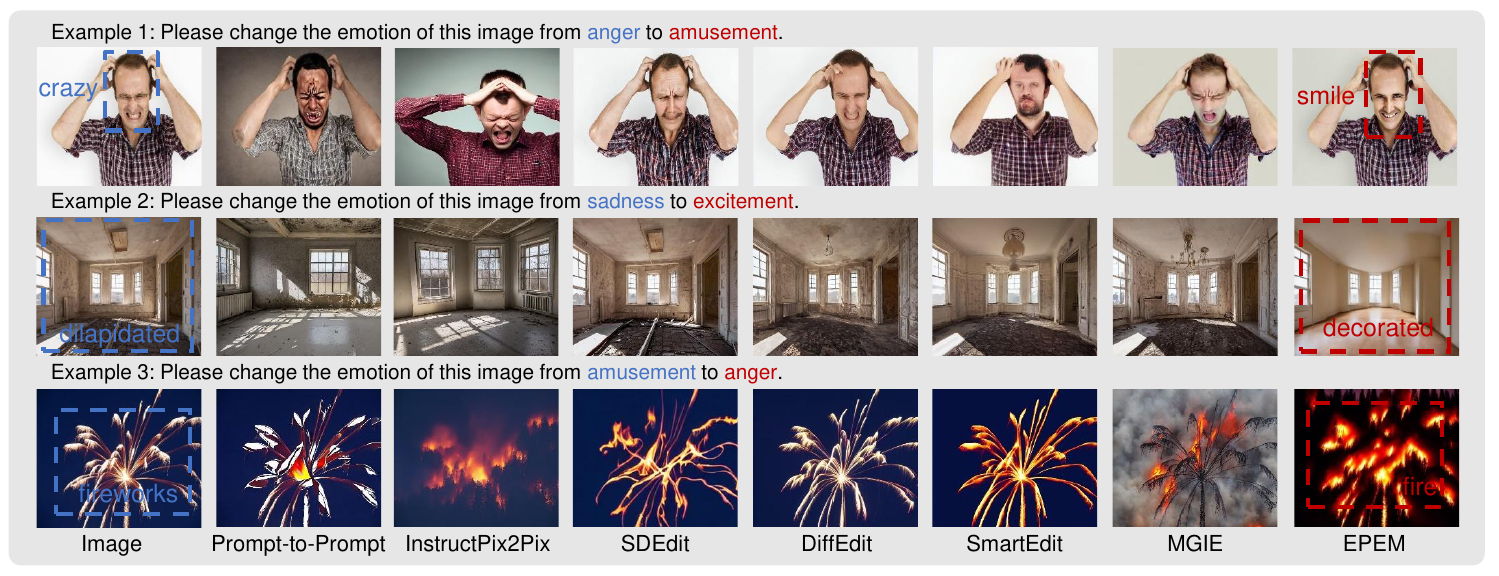}
    \setlength{\abovecaptionskip}{-2.5 ex}
    \setlength{\belowcaptionskip}{-2.5 ex}
\caption{Qualitative comparison of our EPEM approach and several advanced visual customization models. The boxes and texts in blue and red indicate the visual elements before and after editing via our EPEM approach.}
\label{fig:case}
\end{center}
\end{figure*} 

\subsection{Are Emotion-agnostic Contents Retaining?}
To validate the ability of our EPEM approach to retain the emotion-agnostic contents inside images (i.e., exter-emotion semantic retaining challenge), we conduct quantitative and qualitative experiments on our L-AVC dataset. 
\textbf{(1) Quantitative} results are shown in Table \ref{tab:ablation_results}. From this table, we can see that our EPEM approach surpasses the best results underlined by an average improvement 2.92\% ($p$-value<0.05) and 3.21\% ($p$-value<0.05) in FID and CLIP-I on five visual elements. This indicates that the edited images generated by EPEM are close to original image, justifying that EPEM can retain the original contents of images. 
\textbf{(2) Qualitative} analysis is shown in Figure \ref{fig:analysis}. From this figure, we can intuitively observe that EPEM without EAI block (f) changes the emotion of the original image, from delicious fruit cake (\emph{amusement}) to immature and sour strawberries (\emph{disgust}). Our EPEM approach does not change the emotion-agnostic contents of the fruit cake in the original image. Instead, part of the ingredients in the cake are turned into rotten food to \emph{disgust} emotions. Comparing the three pictures, it is obvious that the image contents of (e) and (g) are more similar. This indicates that EAI block can retain emotion-agnostic contents, and further justifies the effectiveness of our EPEM approach in precisely manipulating subjective emotions. 

\subsection{Contributions of Key Components}
To further investigate the influence of key components within our EPEM approach, we conduct a series of ablation studies as shown in Table \ref{tab:main_results} and Table \ref{tab:ablation_results}. From these tables, we can see that:
\textbf{(1) w/o EIC} module exhibits inferior performance compared to our EPEM approach, with an obvious decrease 8.22\%, 15.4\% and 14.8\% in M-Eval, G-Eval and H-Eval three emotional accuracy evaluation metrics. Statistical significance tests show that these improvements are significant ($p$-value<0.01). This justifies the effectiveness of EIC module in understanding emotion semantic conversion, encouraging us to leverage model editing to align emotion conversion in semantic. Furthermore, the H-Eval of different visual elements also shows inferior performance by an average decrease 17.8\% ($p$-value<0.01) compared to our EPEM approach, which further justifies that EIC module could align the emotion conversion with visual semantics. 
\textbf{(2) w/o EAI} block also displays inferior performance compared to our EPEM approach with an obvious increase 3.4\% ($p$-value<0.05) in FID and a decrease 4.96\% ($p$-value<0.05) on CLIP-I in Table \ref{tab:main_results}. This justifies the effectiveness of EAI block in retaining the consistency of original image contents. Besides, for different visual elements, EPEM surpasses EPEM w/o EAI block with the average improvements by 3.46\% ($p$-value<0.05) and 4.96\% ($p$-value<0.05) on FID and CLIP-I two metrics, which further justifies that EAI block could enhance the quality of MLLM emotional guidance and consequently retain emotion-agnostic contents.

\subsection{Visualization Study}
To further justify the effectiveness of our EPEM approach on the L-AVC task, we provide a visualization analysis as shown in Figure \ref{fig:case}. Specifically, we randomly choose the generated images from our EPEM approach and several advanced visual customization models to compare the intuitive effect of emotion editing. From this figure, we can see that: 
\textbf{(1)} Traditional visual customization models, such as Prompt-to-Prompt and InstructPix2Pix, are difficult to understand emotion editing instructions and change visual elements in the original images as exemplified in Example 1. While the other two models SDEdit and DiffEdit can roughly retain the original image contents, but also difficult to achieve accurate emotion manipulation according to the instructions.
\textbf{(2)} MLLM-assisted visual customization models, i.e., SmartEdit and MGIE, are able to basically understand the visual elements that need to be edited, and retain the others roughly consistent, but it is difficult to achieve emotion changes.
\textbf{(3)} In contrast, our EPEM approach equipping with EIC and PER modules can not only achieve precisely emotion manipulating, but also effectively retain emotion-agnostic contents, which further justifies the effectiveness of our EPEM approach on the L-AVC task.  

\section{Conclusion}
In this paper, we propose an Efficient and Precise Emotion Manipulating (EPEM) approach to handle the LLM-centric affective visual customization (L-AVC) task. Specifically, an Efficient Inter-emotion Converting (EIC) module is tailored to address the inter-emotion semantic conversion challenge in the L-AVC task, followed by a Precise Extra-emotion Retaining (PER) module with a designed Emotion Attention Interaction (EAI) block to address the exter-emotion semantic retaining challenge. To comprehensively evaluate the effectiveness of EPEM, we construct an L-AVC dataset, and extensive experimental results on this dataset demonstrate the superior performance of EPEM over several state-of-the-art baselines. In our future work, we would like to extend the L-AVC task to a wide range of scenarios, such as biased and fake news data, aiming to avoid the generation of harmful images. In addition, we would like to manipulate emotions in video, like Sora~\cite{Sora}, aiming to control the generation of harmful emotion videos.

\begin{acks}
We sincerely thank our anonymous reviewers for their helpful comments. This work was supported by two NSFC grants, i.e., No.62576234, No.62376178 and sponsored by CIPS-LMG Huawei Innovation Fund. This work was also supported by Collaborative Innovation Center of Novel Software Technology and Industrialization, and a Project Funded by the Priority Academic Program Development of Jiangsu Higher Education Institutions (PAPD).
\end{acks}

\bibliographystyle{ACM-Reference-Format}










\end{document}